\documentclass[conference]{IEEEtran}
\IEEEoverridecommandlockouts
\usepackage{cite}
\usepackage{amsmath,amssymb,amsfonts}
\usepackage{algorithmic}
\usepackage{graphicx}
\usepackage{textcomp}
\usepackage{xcolor}
\def\BibTeX{{\rm B\kern-.05em{\sc i\kern-.025em b}\kern-.08em
    T\kern-.1667em\lower.7ex\hbox{E}\kern-.125emX}}

\begin{document}

\title{A Factorization Approach for Motor Imagery Classification\\

\thanks{This work was partly supported by Institute of Information \& Communications Technology Planning \& Evaluation (IITP) grant funded by the Korea government (MSIT) (No. 2017-0-00432, Development of Non-Invasive Integrated BCI SW Platform to Control Home Appliances and External Devices by User’s Thought via AR/VR Interface; No. 2017-0-00451, Development of BCI based Brain and Cognitive Computing Technology for Recognizing User’s Intentions using Deep Learning; No. 2019-0-00079, Artificial Intelligence Graduate School Program, Korea University).}
}

\author{\IEEEauthorblockN{Byeong-Hoo Lee}
\IEEEauthorblockA{\textit{Dept. Brain and Cognitive Engineering} \\
\textit{Korea University}\\
Seoul, Republic of Korea \\
bh\_lee@korea.ac.kr}
\and
\IEEEauthorblockN{Jeong-Hyun Cho}
\IEEEauthorblockA{\textit{Dept. Brain and Cognitive Engineering} \\
\textit{Korea University}\\
Seoul, Republic of Korea \\
jh\_cho@korea.ac.kr}
\and
\IEEEauthorblockN{Byoung-Hee Kwon}
\IEEEauthorblockA{\textit{Dept. Brain and Cognitive Engineering} \\
\textit{Korea University}\\
Seoul, Republic of Korea \\
bh\_kwon@korea.ac.kr}

}

\maketitle

% As a general rule, do not put math, special symbols or citations
% in the abstract
\begin{abstract}
Brain-computer interface uses brain signals to communicate with external devices without actual control. Many studies have been conducted to classify motor imagery based on machine learning. However, classifying imagery data with sparse spatial characteristics, such as single-arm motor imagery, remains a challenge. In this paper, we proposed a method to factorize EEG signals into two groups to classify motor imagery even if spatial features are sparse. Based on adversarial learning, we focused on extracting common features of EEG signals which are robust to noise and extracting only signal features. In addition, class-specific features were extracted which are specialized for class classification. Finally, the proposed method classifies the classes by representing the features of the two groups as one embedding space. Through experiments, we confirmed the feasibility that extracting features into two groups is advantageous for datasets that contain sparse spatial features.
\end{abstract}

\begin{small}
\end{small}

\IEEEpeerreviewmaketitle

\section{Introduction}
Recently, advanced machine learning have proven their capabilities in many fields such as computer vision \cite{liu2017quality, yue2021counterfactual}, speech processing \cite{Lee_Yoon_Noh_Kim_Lee_2021, yoon20_interspeech, kim2021fre,lee2021voicemixer}, and bio-signal processing \cite{zhang2019strength, zhang2017hybrid, lee2018high, won2015effect}. Despite these technological advances, there are several challenges in dealing with continuous signals. The absence of ground truth, low signal-noise ratio (SNR), and oscillatory waveform are representative challenges. In this study, we select electroencephalography (EEG) which is one of the mainly used brain signals of non-invasive brain-computer interfaces (BCIs) to classify user intention \cite{nonstationary, MI2}. BCI utilizes brain signals to control and communicate with the external devices without control \cite{bci2, bci3}. Non-invasive BCI is commonly studied because it does not require brain surgery \cite{lee2015subject}. Although EEG signals include user's movements, speech, and even its intention, decoding EEG signals is difficult since the challenges mentioned above also exist in the EEG signals \cite{lee2020continuous,wheelchair}. 

\begin{figure*}[!t]
  \centerline{\includegraphics[scale = 1.2 ]{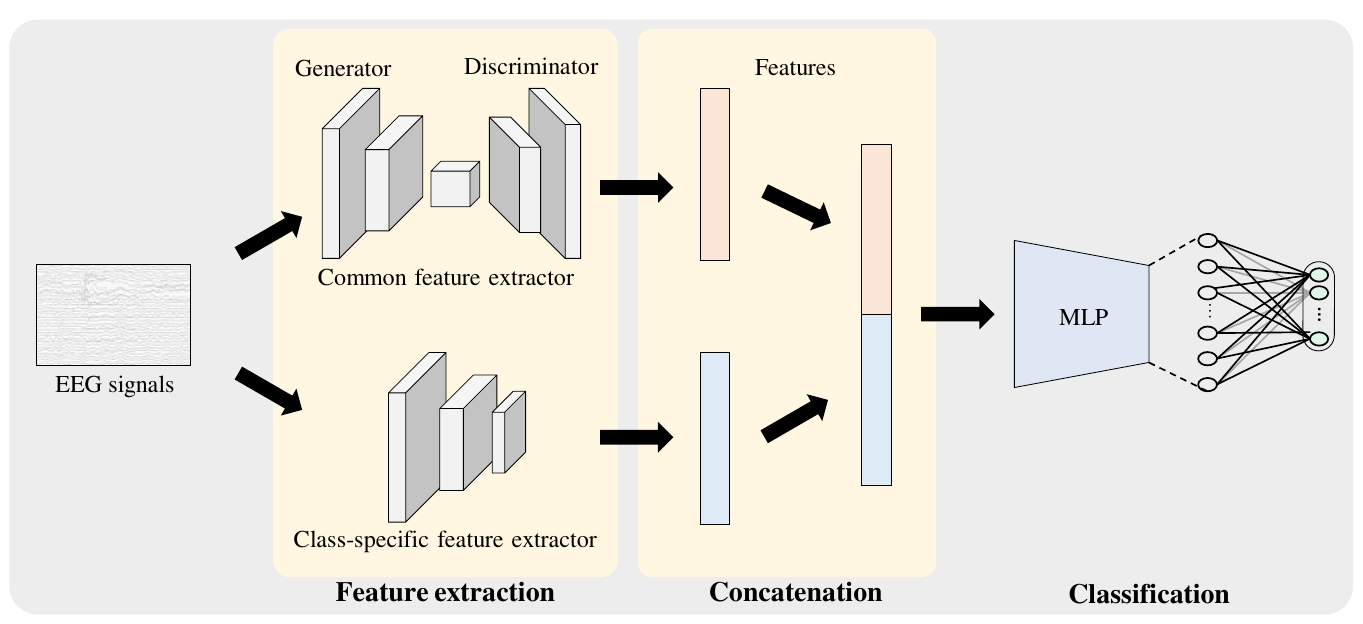}}
  \caption{Flowchart of the proposed method. EEG signals fed into a common feature extractor and a class-specific feature extractor. The generator of the common feature extractor is trained to acquire signal features based on adversarial learning. The class-specific feature extractor is composed of convolutional layers to extract features for classification. Features generated from the two extractors are concatenated and then sent to the multilayer perceptron (MLP). Finally, the MLP proceeds with the classification by representing two features in one feature space.}
\end{figure*}

To generate EEG signals, three endogenous paradigms are commonly selected: motor imagery \cite{OYKwon, JHJeong, IntuitiveMI}, speech imagery \cite{DaSalla, Nguyen}, and visual imagery \cite{sousa2017pure, kwon2020decoding}. Numerous studies have shown that machine learning-based decoding methods are effective in decoding EEG signals. For motor imagery classification, Ang \textit{et al.} \cite{FBCSP} proposed filter bank common spatial pattern (FBCSP) to select key temporal-spatial features for classification by selecting optimal filters from bank. Inspired by FBCSP, Schirrmeister \textit{et al.} \cite{deepconvnet} designed convolutional neural network (CNN) with three different depths. The proposed shallow CNN specialized in motor imagery classification by extracting band power features. Depth-wise convolution and separable convolution were introduced to motor imagery classification in Lawhern \textit{et al.} \cite{eegnet}. Amin \textit{et al.} \cite{MCNN} proposed different depths of CNN for multi-view of feature extraction. Accordingly, their model augments training data at the same time, extracts various features. For speech imagery classification, DaSalla $et \ al.$\cite{DaSalla} proved that spatial pattern (CSP) is effective in speech imagery, like motor imagery. Torres-Garcia $et\ al.$ \cite{Torres-Garcia} proposed fuzzy inference to select optimal channels by finding Pareto front. Nguyen $et\ al.$ \cite{Nguyen} designed Riemannian manifold to decode long and short words. They showed that meaning and word complexity affect decoding performance. Recently, visual imagery induces control signals in a way that visually imagines movement and motion, named visual motion imagery \cite{sousa2017pure}. Sousa \textit{et al.} \cite{sousa2017pure} presented the feasibility of visual motion imagery by comparing it with motor imagery. Kwon \textit{et al.} \cite{kwon2020decoding} used functional connectivity to make the most of the spatial features of the EEG signals. Furthermore, they proved that the three-dimensional visual motion imagery training platform helps users generate EEG signals more consistently and clearly.

In this paper, we proposed a factorization approach to extract common EEG features and class-specific features based on adversarial learning. We assume that features can be divided into two groups. The first group is composed of common features of the EEG signals. Common features contain signal information regardless of classes. Therefore, the influence of noise on feature extraction is reduced. As adversarial learning, the generator learns to extract the common EEG features regardless of the class while deceiving the discriminator. The other group is class-specific features extracted for class classification. The decoding network conducts classification using the two groups of features.

\begin{table}[!t]
{\normalsize
\caption{Architecture of The Proposed Method}
\renewcommand{\arraystretch}{1.2}
\resizebox{\columnwidth}{!}{%
\begin{tabular}{c|cc|cc} \hline
                       & \textbf{Generator} & \textbf{Discriminator} & \textbf{Class-Specific Ext.} & \textbf{MLP}       \\ \hline
\textbf{Inputs}        & 22, 1001           & 1, 2560                & 22, 1001                     & 1, 5120            \\\hline
\textbf{}              & Conv2D(1,40)       & Linear(2560, 1280)     & Conv2D(1,40)                 & Linear(5120, 2560) \\
\textbf{Layers}        & Conv2D(40,40)      & Linear(1280, 1280)     & Conv2D(40,40)                & Linear(2560, 1280) \\
\textbf{}              & AvgPool(1,68)      & Linear(1280, 4)        & AvgPool(1,68)                & Linear(1280, 4)    \\\hline
\textbf{Activation}    & \multicolumn{2}{c|}{ELU \cite{ELU}}                     & \multicolumn{2}{c}{ELU \cite{ELU}}                           \\\hline
\textbf{Optimizer}     & \multicolumn{2}{c|}{SGD \cite{bottou2010large}}                     & \multicolumn{2}{c}{SGD \cite{bottou2010large}}                           \\\hline
\textbf{Loss Function} & \multicolumn{2}{c|}{Adversarial Loss \cite{goodfellow2020generative}}   & \multicolumn{2}{c}{Cross Entropy Loss \cite{shore1980axiomatic}}                     \\\hline
\end{tabular}}}
\end{table}

\section{Methods}

We propose a framework that combines adversarial learning and convolutional neural networks (CNNs). Two identical CNN architectures are designed to factorize EEG signals. One CNN learns to extract class-specific features through the conventional training method. The other CNN is a generator that extracts common features of all EEG signals in the dataset through adversarial learning. Therefore, it is unlikely that features extracted from noise are included. The features are combined with class-specific features, and the classifier predicts the classes based on richer features. The overall flow chart of the proposed framework is depicted in Fig. 1. 
 
\subsection{Common Feature Extractor}
We designed a generator based on convolutional layers and ELU \cite{ELU} activation function. The generator generates features through 2-D convolution filters and average pooling. The final result is 1-D features, and the discriminator distinguishes whether the features are fake or not. A discriminator was designed using linear regression layers and ELU activation function. For adversarial learning, we assigned labels of real and fake to EEG signals and artificially created random noise, respectively. Thus, the generator uses EEG signals for training regardless of the class. As training progresses, the generator extracts common features of EEG signals in the training dataset. Therefore, the generator learns to focus on the signal characteristics of the EEG signals, and the effect of noise on feature extraction is reduced. 

\subsection{Class-Specific Feature Extractor}
Contrary to the common feature extractor, a class-specific feature extractor is designed to extract features for classification. It has an identical structure as the generator of the common feature extractor. Class labels are used for training, and weights are updated based on gradients that are generated from classification errors. Outputs are 1-D features that have the same size as the features from the common feature extractor. 

\subsection{Multilayer Perceptron}
The two features are combined into a feature axis, and MLP represents the combined features as an embedding space. Therefore, MLP is designed to embed the factorized features into a single feature space and proceed with classification based on it. We designed the MLP using linear regression layers and ELU activation function. The details of architectures are described in Table I.

\subsection{Loss Function}
The proposed framework generates two loss functions. The loss function consists of the sum of the two loss functions which are adversarial loss $L_{adv}$ and cross entropy loss $L_{ce}$, defined as

\begin{equation}
loss = L_{adv} + L_{ce}
\end{equation}
Regularization term for the $L_{adv}$ is not applied in this study. Details of $L_{adv}$ is described as follows: 

\begin{equation}
    \begin{split}
    min_G max_D V(D,G) = E_{x \sim P_{data(x)}}[logD(x)] \\
    + E_{z \sim P{_z(z)}}[log(1 - D(G(z))]
    \end{split}
\end{equation}
where $x \sim P_{data}(x)$ is a distribution of real data and $z \sim P_x(x)$ is a latent distribution (e.g., normal distribution). The first loss function is the expected value obtained when the result of putting real data $x$ into the discriminator is log taken. The second term is the expected value obtained when the fake data $z$ is put into the generator and the result is log (1-result) when the result is put into the discriminator. 

$L_{ce}$ is cross entropy loss function for multi class classification, defined as

\begin{equation}
L_{ce} = \sum_{1}^{N}y\log{\bar{y}}
\end{equation}
where $N$, $y$, and $\bar{y}$ are the number of class, true label, and predicted label, respectively. 

\section{Results and Discussions}
We used single-arm movements motor imagery dataset \cite{jeong2020multimodal} that consists of 50 trials for each class. We comprised dataset as 4-class of horizontal reaching (forward, backward, left, and right) and vertical reaching classes (up, down, left, right) among 12-class in session 1. Since the dataset is not divided into a training set and a test set, we randomly selected 10 samples for a class as a test set. Therefore, it consisted of 40 training sets and 10 test sets per class, and 10-fold cross-validation was applied to the training set. Learning rate was 0.0001, iteration patience was 5, batch size was 32, training epoch was 200 epochs, and adamW \cite{AdamW} optimizer with weight decay (0.01) for this experiment. Experimental environment was conducted on an Intel 3.60 Core i7 9700 K CPU with 32 GB of RAM, two NVIDIA TITAN V GPU, CUDA/Cudnn, and Python version 3.9 with PyTorch version 1.9. 

\begin{table}[!t]
{\normalsize
\caption{Comparison of classification accuracy. Reported results are fold-averaged accuracy. The upper subscripts $^1$ and $^2$ denote deep and shallow ConvNet \cite{deepconvnet}, respectively. RF denotes random forest. The highest numbers are bold.}
\renewcommand{\arraystretch}{1}
\begin{tabular}{ccc} \hline
\multicolumn{1}{l}{} & \textbf{Horizontal}     & \textbf{Vertical}       \\
\textbf{Model}       & \textbf{Accuracy (std)} & \textbf{Accuracy (std)} \\\hline
CSP+RF \cite{qi2012random}       & 29.82 (3.52)            & 33.16 (4.47)            \\
CSP+SVM \cite{FBCSP}             & 29.11 (2.97)            & 32.89 (4.34)            \\
CSP+LDA \cite{FBCSP}             & 33.35 (3.44)            & 38.14 (3.19)            \\
FBCSP \cite{FBCSP}               & 41.57 (\textbf{4.40})            & 46.23 (4.54)            \\
EEGNet \cite{eegnet}              & 48.39 (2.98)            & 52.76 (2.42)            \\
ConvNet$^1$ \cite{deepconvnet}     & 47.34 (2.77)            & 51.48 (\textbf{5.93})            \\
ConvNet$^2$ \cite{deepconvnet}     & 51.34 (2.77)            & 55.67 (3.69)            \\
Proposed Method      & \textbf{54.29} (3.40)            & \textbf{57.29} (5.30)           \\\hline
\end{tabular}}
\end{table}

Accuracy was calculated as the average of the accuracy of all folds for the test set. The common spatial pattern (CSP) algorithm is a useful method for motor imagery classification. CSP is a method of maximizing the variance of one class and minimizing the variance of the other based on the covariance matrix. Therefore, when the number of data is relatively small, it occurs overfitting and performance decrease. Random forest, support vector machine, and linear discriminant analysis were selected as a classifier for CSP. According to Table II, CSP showed relatively low accuracy regardless of the classifier. Overfitting due to the small number of data used in this study can be one of the causes of performance degradation. Another possibility is that the composition of the class is movements within a single-arm, thus it is a limit to extracting spatial features. To alleviate the overfitting, the FBCSP, which combines the filter bank and the CSP, recorded about an 8\% improvement in accuracy compared to the conventional CSP. 

CNN-based deep learning has also been actively studied for motor imagery classification, and in this paper, three CNN methods \cite{deepconvnet, eegnet} which are publicly opened were used. EEGNet \cite{eegnet} and deep ConvNet \cite{deepconvnet} showed similar performance in both horizontal and vertical reaching classification. However, shallow ConvNet showed 4\% higher accuracy than deep ConvNet. Although shallow ConvNet has a shallow architecture, it extracts band power features, which are known as key features of motion imagery. The proposed method focuses on extracting richer features by factorizing the EEG signals than the existing methods. In particular, since the composition of the data has limitations to consider spatial features, the existing feature extraction method does not seem to maximize performance. Therefore, in the case of sparse spatial features, it is advantageous to consider both the common characteristics of the signals and the class-specific features according to Table II.

However, it has not yet been proven whether the common features are robust to signal noise and whether the EEG signal was factorized in a proper direction for motor imagery classification. In particular, it is challenging to obtain the ground truth of clean EEG signals. Hence study without ground-truth prevents deriving a general solution, results should be thoroughly verified. Therefore, our future research will proceed in the direction of solving the challenges mentioned above. 
\section{Conclusion and Future Works}
\label{sec:print}
In this paper, we proposed a factorization approach that extracts common EEG features and class-specific features. The common EEG features are the features of all EEG signals existing in the dataset. they are assumed to be extracted. Therefore it is assumed that these features are robust to signal noise. On the other hand, class-specific features are specialized in class classification, although they are prone to be affected by signal noise. According to the results, when classifying data with sparse spatial features, such as single-arm movements imagery, the features factorized into two groups are more advantageous than the generally extracted features. However, since there are still challenges to be proved, our future works will proceed in the direction of solving them.

\section{Acknowledgement}
The author would like to thank H.-H. Park and H.-B. Shin for help with the discussion of the data analysis.\\
\bibliographystyle{IEEEtran}
\bibliography{Reference}

% that's all folks
\end{document}